\documentclass[11pt]{article}

\usepackage[preprint]{acl}

\usepackage{times}
\usepackage{latexsym}

\usepackage[T1]{fontenc}

\usepackage[utf8]{inputenc}

\usepackage{microtype}

\usepackage{amsmath}
\usepackage{amssymb}
\usepackage{tabularx}

\usepackage{inconsolata}

\usepackage{graphicx}

\usepackage{booktabs, multirow}

\usepackage{tikz}
\usetikzlibrary{positioning, arrows.meta, fit, backgrounds, calc, shapes.geometric}
\usepackage{xcolor}
\definecolor{iceTeal}{HTML}{1B9E8A}
\definecolor{iceOrange}{HTML}{E07B39}
\definecolor{iceGrey}{HTML}{8A8A8A}
\definecolor{iceInk}{HTML}{2B2B2B}

\title{ Instruction-Conditioned Exploration for Reinforcement Learning \\ with Self-Distillation to an Unconditioned Policy}

\author{Jim Dilkes \\
  {University of Southampton} \\
  Southampton \\
  United Kingdom \\
  \texttt{j.dilkes@soton.ac.uk} \\\And
  Vahid Yazdanpanah \\
  {University of Southampton} \\
  Southampton \\
  United Kingdom \\
  \texttt{V.Yazdanpanah@soton.ac.uk} \\\And
  Sebastian Stein \\
  {University of Southampton} \\
  Southampton \\
  United Kingdom \\
  \texttt{ss2@ecs.soton.ac.uk}
}

\begin{document}
\maketitle
\begin{abstract}
Post-training Large Language Models (LLMs) with Reinforcement Learning (RL) has become an important tool for improving model capabilities, but the LLM action-space structure introduces challenges distinct from classical RL, with implications for inducing exploration. New methods are required that leverage the broad knowledge and flexibility of pre-trained LLMs to deliberately generate diverse experience at training time. We propose Instruction-Conditioned Exploration (ICE), which appends one of a small fixed set of instructions to task prompts during training, using the same set for every problem, increasing the coverage of behaviours attempted. To facilitate ICE, we combine RL on the instruction-conditioned policy with self-distillation of its correct rollouts into the unconditioned test-time policy. 
ICE with this objective improves Qwen3-1.7B held-out pass@1 performance at 4K response length on mathematical reasoning tasks by $5.0\%$ relative to training with DAPO, with improvement persisting at a longer 8K context. The improvement does not appear for Qwen3-4B at 4K, where the instructions do not expand base-model coverage.
\end{abstract}

\section{Introduction}
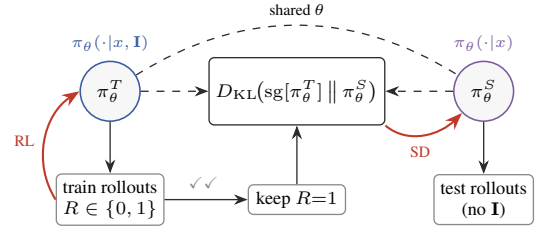
\begin{figure}[t]
\centering
\definecolor{iceTeacher}{HTML}{3A5BA0}
\definecolor{iceStudent}{HTML}{7A5C9E}
\definecolor{iceRed}{HTML}{C0392B}
\resizebox{\columnwidth}{!}{%
\begin{tikzpicture}[
  font=\scriptsize,
  >=Stealth,
  box/.style={rounded corners=2pt, draw=iceInk, line width=0.5pt, inner sep=2.5pt, align=center, fill=white},
  policy/.style={circle, draw=iceInk, line width=0.6pt, fill=iceInk!4, minimum size=8mm, align=center, inner sep=0pt},
  arr/.style={->, line width=0.5pt, draw=iceInk},
  parr/.style={->, line width=0.8pt, draw=iceRed},
  darr/.style={dashed, line width=0.5pt, draw=iceInk},
  hdr/.style={font=\footnotesize\sffamily\bfseries, text=iceInk},
]
\node[hdr, anchor=west] at (-0.2,2.05) {1. Train-time exploration (ICE)};
\node[box] (px)  at (0,0.85) {$x$};
\node[box, draw=iceTeal]   (pi1) at (0,0.20) {$x + \mathbf{I}_1$};
\node[box, draw=iceOrange] (pi2) at (0,-0.45) {$x + \mathbf{I}_2$};
\node[font=\tiny] (pdots) at (0,-0.85) {$\vdots$};
\node[box, draw=iceGrey]   (piN) at (0,-1.25) {$x + \mathbf{I}_N$};
\node[policy] (pol) at (2.5,-0.2) {$\pi_\theta$};
\draw[arr] (px)  -- (pol);
\draw[arr] (pi1) -- (pol);
\draw[arr] (pi2) -- (pol);
\draw[arr] (piN) -- (pol);
\begin{scope}[shift={(5.4,-0.2)}]
  \draw[densely dashed, iceInk, line width=0.6pt] (0,0) ellipse (1.5 and 1.25);
  \begin{scope}[on background layer]
    \fill[iceGrey,   opacity=0.30] (0.05,-0.15) ellipse (0.55 and 0.45);
    \fill[iceTeal,   opacity=0.30] (-0.5,0.35)  ellipse (0.6 and 0.5);
    \fill[iceOrange, opacity=0.30] (0.55,0.4)   ellipse (0.6 and 0.5);
    \fill[iceGrey,   opacity=0.30] (0.45,-0.55) ellipse (0.55 and 0.45);
  \end{scope}
  \node[font=\tiny\itshape, iceGrey] at (0.05,-0.15) {$x$ alone};
  \node[font=\tiny, text=iceInk, align=center, anchor=north] at (0,-1.35)
     {$\mathrm{supp}\,\pi_\theta^{\mathrm{comp}}=\bigcup_{\mathbf{I}}\mathrm{supp}\,\pi(\cdot|x,\mathbf{I})$};
\end{scope}
\draw[arr] (pol) -- (3.85,-0.2);
\node[font=\tiny, anchor=south, align=center] at (4.7,0.95) {broadened\\support};
\draw[iceGrey!50, line width=0.4pt] (-0.5,-2.15) -- (7.0,-2.15);
\begin{scope}[shift={(0,-4.25)}]
\node[hdr, anchor=west] at (-0.2,1.7) {2. Behaviour transfer (RL + filtered SD)};
\node[policy, draw=iceTeacher] (teach) at (1.4,0.05) {$\pi_\theta^{T}$};
\node[font=\tiny, anchor=south, text=iceTeacher] at (1.4,0.45) {$\pi_\theta(\cdot|x,\mathbf{I})$};
\node[box, draw=iceGrey, minimum width=13mm] (roll) at (1.4,-1.4) {train rollouts\\$R\in\{0,1\}$};
\draw[arr] (teach) -- (roll);
\draw[parr] (roll.west) to[bend left=45] node[left, font=\tiny, text=iceRed] {RL} (teach.west);
\node[box, draw=iceInk, minimum width=20mm, minimum height=9mm] (kl) at (3.9,0.05)
  {$D_{\mathrm{KL}}\!\big(\mathrm{sg}[\pi_\theta^{T}]\,\big\|\,\pi_\theta^{S}\big)$};
\node[box, draw=iceGrey, minimum width=13mm] (filt) at (3.9,-1.4) {keep $R{=}1$};
\draw[arr] (roll.east) -- (filt.west);
\node[font=\tiny, text=iceGrey, anchor=south] at (2.65,-1.38) {$\checkmark\checkmark$};
\draw[arr] (filt.north) -- (kl.south);
\draw[darr, ->] (teach.east) -- (kl.west);
\node[policy, draw=iceStudent] (stud) at (6.4,0.05) {$\pi_\theta^{S}$};
\node[font=\tiny, anchor=south, text=iceStudent] at (6.4,0.45) {$\pi_\theta(\cdot|x)$};
\node[box, draw=iceGrey, minimum width=13mm] (test) at (6.4,-1.4) {test rollouts\\(no $\mathbf{I}$)};
\draw[darr, ->] (stud.west) -- (kl.east);
\draw[parr] (kl.south east) to[out=-40, in=225] node[below, font=\tiny, text=iceRed, pos=0.45] {SD} (stud.225);
\draw[arr] (stud) -- (test);
\draw[darr] (teach) to[bend left=32] node[above, font=\tiny, pos=0.5] {shared $\theta$} (stud);
\end{scope}
\end{tikzpicture}%
}
\caption{Overview. \textbf{(1)} ICE supplements the task prompt $x$ with one of several behavioural instructions $\mathbf{I}$ per rollout, broadening the support available to explore at training time. \textbf{(2)} Reinforcement learning optimises the instruction-conditioned teacher rollouts, with simultaneous self-distillation of the correct teacher trajectories into the unconditioned student, which is the only policy used at test time.}
\label{fig:overview}
\end{figure}

Post-training with Reinforcement Learning (RL) is increasingly used to improve and tune Large Language Models (LLMs), including for general-purpose agents \citep{qi_webrl_2025} and deployed real-world systems \citep{jiang_improving_2025}.
Underlying these applications is a need for sustained exploration of diverse behaviours during training: on-policy RL provides direct learning signal only for sampled actions; behaviours never produced during training cannot be directly reinforced. Classical RL exploration methods rely on stochasticity in action selection (e.g.\ $\epsilon$-greedy exploration or action-noise injection), assuming an action space structured such that random perturbation produces useful candidate trajectories \citep{sutton_reinforcement_2018}. 

In LLMs this assumption fails: ``actions'' correspond to tokens drawn from vocabularies of tens of thousands, and tokens must be combined in highly specific sequences to be meaningful, so random token-level perturbation lands off the manifold of coherent text rather than producing novel viable trajectories. Nor does restricting sampling to probable tokens resolve this, as superficially diverse text can carry the same meaning \citep{wang_ragen-2_2026}; behavioural diversity requires exploration above the token level. Compounding the problem, RL post-training methods are known to sharpen rather than broaden the model's distribution \citep{yue_does_2025}, motivating new approaches to increase the coverage of behaviours experienced during training, what \citet{wu_invisible_2026} call \textit{support expansion}.

Existing exploration methods operate at the token or sequence level \citep{cui_entropy_2025, he_rewarding_2025}, with no mechanism for surfacing low-probability behaviours, or use context mediation \citep{lu_prompt_2026, szot_expanding_2026} that either relies on capability-bounded self-feedback or lacks an explicit mechanism for transferring train-time behaviours to test time.

To address these limitations, we introduce \textbf{Instruction-Conditioned Exploration} (ICE) and train it with a combination of RL and filtered self-distillation (Figure~\ref{fig:overview}). {ICE} is our novel approach to generating diverse training samples: at training time, we condition text generation on instructions sampled from a set of task strategies, inducing diverse rollouts from this \textit{instruction-conditioned} (teacher) policy. The same model, queried without an appended instruction, defines an \textit{unconditioned} (student) policy. 

The training objective combines two components on a shared set of teacher rollouts. First, a reward-maximising RL term updates the teacher policy (we use the DAPO \citep{yu_dapo_2025} variant of the GRPO RL algorithm \citep{shao_deepseekmath_2024}). Second, a forward-KL distillation term updates the student policy, pulling its generation distribution toward the teacher rollouts that received a correct answer, so that at test time the unconditioned policy replicates useful behaviours that the teacher samples during training. 

The diverse instructions target support expansion directly rather than relying on capability-bounded self-feedback, and the distillation term supplies the explicit transfer to the unconditioned policy that prompt-diversity methods leave to parameter sharing. Because we filter to correct rollouts, the instructions need not be helpful on every training example. We require only a small set of generic instructions covering a diverse range of potentially useful behaviours, a weaker requirement than that of privileged-information methods, where the added context is helpful to each specific problem by construction.

Empirically, on Qwen3 \citep{yang_qwen3_2025} at 1.7B parameters and 4K response length, ICE with the filtered objective improves the macro-mean pass@1 across held-out mathematical reasoning benchmarks by $5.0\%$ relative to DAPO, with 5/5 benchmark wins and a 95\% bootstrap CI excluding zero ($n=5$). The improvement over ICE with RL alone is smaller and less certain. A smaller improvement persists at the longer-context 1.7B/8K setting, while at 4B/4K it does not improve over DAPO. This is consistent with the instructions not expanding base-model pass@256 coverage at 4B. Comparing base-model pass@$k$ with and without instructions therefore serves as a cheap test of where ICE can help. 

Our contributions are as follows:
\begin{itemize}
    \item A novel method of instruction conditioning to induce diverse rollouts from LLMs, driving exploration during RL post-training (ICE).
    \item The combination of correctness-filtered self-distillation with a diverse train-time prompting approach, enabling internalisation of only the useful behaviours induced by a set of distinct instructions but without requiring all individual instructions to be helpful.
    \item A multi-seed empirical study at various model scales, context lengths and self-distillation settings, demonstrating the improvement of ICE with the filtered objective over a DAPO baseline at the 1.7B parameter scale.
\end{itemize}

\section{Related Work}

\subsection{Exploration in RL Post-Training of LLMs}
\label{sec:related_exploration_llms}
A considerable focus of recent work on LLM post-training is using RL to improve performance on tasks with verifiable outcomes (RLVR). Learning under RL requires actions producing viable trajectories to be sampled with non-negligible probability during training, motivating directed exploration rather than the stochastic methods sufficient in classical RL. 

This need is reinforced by recent findings that RL post-training primarily sharpens the distribution over outputs already accessible to the base model, raising pass@1 but reducing pass@$k$ coverage at large $k$ and so narrowing the reasoning boundary \citep{yue_does_2025, zhang_interplay_2025}. \citet{wu_invisible_2026} also find that RLVR reduces the support of the response distribution, excluding previously available correct answers (\textit{support shrinkage}), and call for new methods that instead achieve \textit{support expansion}, up-weighting previously low-probability completions.

Token-level and sequence-level diversity interventions \citep{cui_entropy_2025, wang_beyond_2025, shen_entropy_2025, chen_passk_2025, he_rewarding_2025, zhang_dpepo_2026, song_outcome-based_2025, hu_diversity-incentivized_2026, liang_can_2025} show promise in preventing entropy collapse but do not directly address the need for support expansion: superficial token diversity can mask semantic similarity (what \citet{wang_ragen-2_2026} call \textit{template collapse}).

\subsection{Context-Mediated Exploration in LLMs}
\label{sec:related_context_explore}
Context-mediated exploration methods modify the context provided to the model during training, aiming to achieve support expansion relative to a singular context, through pre-defined hints, self-feedback, or pre-determined templates.

Pre-defined hint based methods \citep{zhou_breaking_2026, zhang_scaf-grpo_2026, liao_self-hinting_2026, chen_nudging_2025} selectively provide the LLM with additional information about how to approach a problem, but require privileged information from expert annotators, more powerful models, or knowledge of the ground-truth answer, limiting scalability.

Self-feedback approaches \citep{madaan_self-refine_2023, bai_constitutional_2022, shinn_reflexion_2023, kim_language_2023, kumar_training_2024, li_lanpo_2025, szot_expanding_2026} let the model make multiple attempts, with previous attempts fed back for critique, refinement, or as demonstrations of failure. Although not typically framed as exploration techniques, these repeated attempts under different contexts broaden the strategies the model produces, achieving support expansion. Their common limitation is reliance on attempt-level self-feedback, which increases rollout cost, remains bounded by current capability, and offers no explicit control over the diversity of explored strategies.

Template approaches use a fixed set of pre-determined templates, each conditioning the task description with a different context during training and directly driving support expansion relative to the unconditioned policy. This is exemplified by \textit{Prompt Augmentation} \citep{lu_prompt_2026}, which conditions each training example on one of a set of response-format templates. However, this introduces a train/test distributional mismatch: the model is trained on various templated prompts but evaluated on a single test-time prompt.

In concurrent work, \citet{lee_nudging_2026} condition RL rollouts on strategy contexts generated per problem by a stronger model, combine this with a modified group advantage, and transfer the conditioned behaviour through an advantage-weighted term rather than a divergence. They report improved mathematical reasoning, outperforming GRPO baselines. 

\subsection{Self-Distillation in LLM Post-Training}
\label{sec:related_selfdistill}
We build on self-distillation, a form of knowledge distillation \citep{hinton_distilling_2015} in which teacher and student are the same model differing in input context, with the teacher typically conditioned on privileged train-time information. 

In early work on distillation for LLMs, \citet{askell_general_2021} improve alignment through in-context learning, then use ``context distillation'' to transfer the behaviour induced by that context into models queried without it, including the original model. The transfer minimises the KL divergence between the output distributions of the conditioned and target models. It matches the alignment performance of the in-context prompt while freeing the context window, which was particularly important in an era of shorter context lengths. \citet{snell_learning_2022} generalise the approach beyond alignment, distilling a teacher prompted with instructions, examples and a scratch-pad into a student that sees only a minimal prompt. In both, the teacher is held fixed and the objective is a divergence alone, with no task reward.

Recent methods reuse this construction and vary along two axes: whether the rollouts used for distillation are teacher-sampled (off-policy) \citep{bhargava_prompt_2024, penaloza_privileged_2026} or student-sampled (on-policy) \citep{zhao_self-distilled_2026, shenfeld_self-distillation_2026, hubotter_reinforcement_2026}, and whether the objective is a divergence alone or also carries a task reward \citep{penaloza_privileged_2026}. Throughout, these approaches differ from standard on-policy distillation, in which the teacher is a separate, stronger model; here the teacher's advantage is contextual information rather than model capability.

We extend $\pi$-distill \citep{penaloza_privileged_2026}, a flexible off-policy method combining teacher and student RL objectives with reverse and forward KL terms, replacing its privileged information with a behavioural instruction $\mathbf{I}$.

\section{Instruction-Conditioned Exploration for LLM RL}

We now introduce Instruction-Conditioned Exploration (ICE), our template-based approach to support expansion in RL training of LLMs, beginning with the DAPO formulation it builds on (\S\ref{sec:preliminary_rl}) before defining the composite instruction-conditioned policy and its RL objective (\S\ref{sec:method_ice}). The test-time transfer problem is addressed separately in \S\ref{sec:method_selfdistill}.

\subsection{Preliminaries: RL for LLMs}
\label{sec:preliminary_rl}
When applying reinforcement learning to LLMs, we treat the model with parameters $\theta$ as a policy $\pi_\theta$ that generates text responses $y$ to input texts $x$ (the task prompt) and receives reward $R(y,x)$. For mathematical reasoning, $x$ contains a question with a verifiable answer; the response $y$ receives reward $1$ if the extracted answer is correct and $0$ otherwise. The RL objective is to maximise ${J}_{\text{RL}}(\theta) = \mathbb{E}_{y \sim \pi_\theta}\left[R(y,x)\right]$ 
over the task distribution \citep{sutton_reinforcement_2018}.

GRPO \citep{shao_deepseekmath_2024} replaces $R$ with a group-normalised advantage $A_i=(R_i-\mu)/\sigma$ over $G$ responses sampled from $\pi_{\theta_\text{old}}$ for the same input $x$, where $\mu,\sigma$ are the mean and standard deviation of the group rewards. DAPO \citep{yu_dapo_2025}, the predominant GRPO variant, adds token-level loss aggregation, asymmetric clipping, dynamic sampling, and overlong reward shaping; we use it as both our baseline and the RL component of our method.

\subsection{Instruction-Conditioned Exploration}
\label{sec:method_ice}

We introduce a new approach to template-based context-mediated exploration in language models, instruction-conditioned exploration (ICE). ICE bears similarity to Prompt Augmentation \citep{lu_prompt_2026}, but supplements task descriptions with opinionated, diverse \textit{behavioural} instructions rather than solution-agnostic formatting instructions. This broadens the support of the generated text distribution relative to a singular task prompt, what we term \textit{support composition} to distinguish it from \textit{support expansion} (defined by \citet{wu_invisible_2026}, see \S\ref{sec:related_exploration_llms}), for which the expansion is due only to parameter changes with respect to the base model. 

Formally, we define a set $\mathcal{I} = \{\mathbf{I}_1, \dots, \mathbf{I}_N\}$ of $N$ instructions that can be independently appended to the problem description during training. Each instruction should contain high-level guidance that pushes the response of the LLM into a distinct direction. For example, for mathematical reasoning problems we might use the instruction: ``\textit{Ask: what symmetry does this problem have, and how can it reduce the number of cases or simplify the expression?}''. If $p(\mathbf{I})$ is the probability that instruction $\mathbf{I}$ will be selected, the composite ICE policy becomes
\begin{equation}
    \pi_\theta^\text{comp}(y|x) = \sum_{\mathbf{I}\in\mathcal{I}} p(\mathbf{I})\pi_\theta(y|x,\mathbf{I}),
\end{equation}
with the support of this policy being the union of the supports of its components
\begin{equation}
    \text{supp}(\pi_\theta^\text{comp}(\cdot|x))=\bigcup_{\mathbf{I}\in\mathcal{I}}\text{supp}(\pi_\theta(\cdot|x,\mathbf{I})).
\end{equation}
We take $\emptyset\in\mathcal{I}$ to denote the unconditioned case having $\pi_\theta(\cdot|x,\emptyset)=\pi_\theta(\cdot|x)$, which is always part of the instruction set so that $\text{supp}(\pi_\theta^\text{comp}(\cdot|x)) \supseteq \text{supp}(\pi_\theta(\cdot|x))$.
The RL objective with ICE is then 
${J}_{\text{ICE}}(\theta) = \mathbb{E}_{y \sim \pi_\theta^\text{comp}}\left[R(y,x)\right]$.

Each rollout receives a single instruction $\mathbf{I}\in\mathcal{I}$ appended to the task prompt. How instructions are distributed across the rollouts of a task is a design choice: they may be sampled per rollout, or allocated so that each group covers several instructions and unconditioned rollouts. The model then generates text conditioned on the augmented prompts. This approach introduces a trade-off: while coverage of different problem-solving strategies may broaden the support of the policy during training, $\mathbf{I}$ could also provide unhelpful instructions, reducing accuracy and therefore positive signal during training. 

We combine this with the GRPO RL objective function, allowing each response to a specific input $x$ to be conditioned on different instructions, thereby creating diversity within each group beyond that of sampling alone.

\section{Self-Distillation of LLMs}
\label{sec:method_selfdistill}
The use of a composite policy at training time requires the transfer of useful behaviours discovered during training to the unconditioned test-time policy. Parameter sharing enables implicit transfer, but an explicit transfer mechanism can strengthen it. We extend the $\pi$-distill off-policy self-distillation objective of \citet{penaloza_privileged_2026} to provide this mechanism, and define the variant used in our main experiments.

In off-policy self-distillation, trajectories sampled from a teacher policy are used to update both the teacher and student policies, with the teacher policy distinguished from the student policy by having additional, privileged information about the task or its solution. In our setting, the privileged information is replaced with a behavioural instruction $\mathbf{I}$.

Let $\pi^T_\theta := \pi_\theta(\cdot \mid x, \mathbf{I})$ and $\pi^S_\theta := \pi_\theta(\cdot \mid x)$ denote the instruction-conditioned (teacher) and unconditioned (student) policies, respectively, sharing parameters $\theta$. The $\pi$-distill objective combines reward-maximising terms for both policies, weighted by $\alpha$, with two KL-divergence terms between the sampled teacher and student policies, weighted by $\beta$. Writing $\text{sg}[\cdot]$ for the stop-gradient operator, $D_{KL}$ for the KL divergence, and $R(y,x)$ for the reward of response $y$ on task $x$, the reward terms are the teacher reward $J_T^R = \mathbb{E}_{y \sim \pi^T_\theta}[R(y,x)]$ and the importance-corrected student reward $J_S^R = \mathbb{E}_{y \sim \pi^T_\theta}[(\pi^S_\theta/\text{sg}[\pi^T_\theta])\, R(y,x)]$, both estimated on teacher rollouts. The two KL terms are a reverse-KL term $J_T^{KL} = -D_{KL}\big(\pi^T_\theta \,\|\, \text{sg}[\pi^S_\theta]\big)$ that pulls the teacher toward the student, and a forward-KL term $J_S^{KL} = -D_{KL}\big(\text{sg}[\pi^T_\theta] \,\|\, \pi^S_\theta\big)$ that pulls the student toward the teacher.

We modify this objective as $J(\theta) = J^R(\theta) + J^{KL}(\theta)$ with components
\begin{equation}
J^R(\theta) = \alpha J_T^R(\theta) + (1-\alpha) J_S^R(\theta)
\end{equation}
\begin{equation}
J^{KL}(\theta) = \beta_T J_T^{KL}(\theta) + \beta_S J_S^{KL}(\theta)
\end{equation}
reusing the reward and KL terms defined above. We retain the constraint on reward weighting to avoid adding overall reward scale as an additional degree of freedom (absorbed by the learning rate). The KL terms do not require such a constraint as they trade off against the reward signal, not against each other. The original $\pi$-distill objective is recovered by setting $\beta_T=\alpha\beta$ and $\beta_S=(1-\alpha)\beta$.

In the case with the student-KL term and no student reward ($\alpha=1$, $\beta_S>0$), we additionally filter teacher-sampled responses to only include those giving a correct answer, $R(y,x)=1$. 
The $\pi$-distill motivation for including the complete, unfiltered KL terms is to keep the importance sampling ratio in the student reward term close to $1$ for all training samples. Without a student reward term, this motivation no longer applies. This is closer to a standard knowledge distillation approach in which we want the student policy to learn to replicate good responses generated by the teacher policy.
The complete objective used in our headline experiments is
\begin{equation}
\label{eqn:rlfsd_objective}
J_{\text{filt}}(\theta) = J_T^R(\theta) - \beta_S\, D_{KL}\big(\text{sg}[\pi^{T+}_\theta] \,\big\|\, \pi^S_\theta\big)
\end{equation}
where $\pi^{T+}_\theta$ is the teacher distribution restricted to correct responses ($R(y,x)=1$, renormalised), so the KL term is $J_S^{KL}$ with $\pi^T_\theta$ replaced by $\pi^{T+}_\theta$, estimated on the correctness-filtered teacher rollouts. $J_T^R$ is optimised with the DAPO objective of \S\ref{sec:preliminary_rl} on the same instruction-conditioned rollouts.

\section{Experimental Setup}
To assess our methods, we apply ICE with modified $\pi$-distill objectives, including the filtered configuration, to mathematical reasoning problems using Qwen3 \citep{yang_qwen3_2025} models with thinking mode disabled, consistent with recent post-training and self-distillation work \citep{lu_prompt_2026, liao_self-hinting_2026, zhao_self-distilled_2026, yang_learning_2026}.
We compare a baseline DAPO configuration to models trained with ICE, comparing variants across selected ($\alpha$, $\beta_T$, $\beta_S$) at 1.7B and 4B parameters, and 4K and 8K (1.7B only due to compute budget constraints) token response lengths. Models are trained for up to 7 epochs of the English language subset of the deduplicated DAPO-Math-17K dataset \citep{yu_dapo_2025} (14.1k examples), ``DAPO-17K-en''. Training was conducted with NVIDIA GH200 superchips on the Isambard-AI cluster \citep{mcintosh-smith_isambard-ai_2024}, using 2 GPUs per seed for 1.7B configurations and 4 GPUs for 4B. Total compute and per-model-size costs are reported in Appendix \ref{apdx:compute}, and artifact licenses in Appendix~\ref{apdx:licenses}.

We select $N=5$ instructions via an automated procedure using frontier models (Claude Sonnet 4.6 and Claude Opus 4.6) to propose candidate strategies and the target model (Qwen3 4B) to select among them by coverage on a held-out 250-problem training subset, yielding the instructions in Table \ref{tab:instructions}. Full details are in Appendix \ref{sec:appendix-instructions}. The selection uses training-set problems, leaving evaluation uncontaminated. In practice, this process could be replaced by human expert instructions.

\paragraph{Algorithm Details and Hyperparameters}
Hyperparameters are not separately tuned; we adopt standard values commonly used in recent RL post-training work \citep{yu_dapo_2025, guo_deepseek-r1_2025, lu_prompt_2026, liao_self-hinting_2026}.
Our implementation builds on the OpenRLHF framework \citep{hu_openrlhf_2024}. To train the models, we used the DAPO algorithm described in \S\ref{sec:preliminary_rl} with rollout group size $G=8$, and clipping bounds $\epsilon_\text{low}=0.2,\epsilon_\text{high}=0.28$.
Rollouts are truncated above $4096$/$8192$ tokens (4K/8K) depending on the setting, with overlong penalty above $3072$/$6144$ tokens respectively. Rollout text is generated with temperature $T=1.0$, $\text{top\_p}=1.0$. Learning rate is $1\text{e}{-6}$, constant after 40 warmup steps. Batch sizes are $128$ for rollout and training, with a training micro-batch of $8$ at 1.7B and $4$ at 4B.
For instruction-conditioned training each of the $N=5$ supplementary instructions is appended to the task prompt for one rollout, and the remaining $3$ rollouts use the task prompt alone. Method variants are summarised in Table \ref{tab:variants}.

\paragraph{Evaluation}
We evaluate the post-trained models on five held-out mathematical reasoning benchmarks: AIME-24 \citep{zhang_american_2024}, AIME-25 \citep{zhang_american_2025}, MATH-500 \citep{hendrycks_measuring_2021}, HMMT-Feb-25, and HMMT-Nov-25 \citep{balunovic_matharena_2026}. MATH-500 contains 500 questions; the AIME and HMMT benchmarks each contain 30 more difficult and more recent questions. All evaluation is performed on the unconditioned policy $\pi_\theta^S$ (defined in \S\ref{sec:method_selfdistill}), without any instructions in the context. At test time we have no mechanism for determining which supplementary instructions will be beneficial; this must be learned by the student policy. 

We additionally evaluate on DAPO-200, a fixed 200-question subset of DAPO-17K-en, which provides a training-set performance measure when models are evaluated unconditioned. 

Evaluation responses are sampled at $T=0.6$, $\text{top\_p}=1.0$, $n=16$ for AIME/HMMT, and $n=1$ for MATH-500/DAPO probe. We use pass@1 (averaged for $n=16$). To identify the best performing evaluation step fairly across runs, we use the DAPO-200 reward to select the best evaluation step of each training run, then report held-out scores at that step. As seed counts are limited, we read results from complementary signals rather than any single statistic: the cross-seed mean$\pm$std, $\Delta$ vs DAPO, per-benchmark wins, and, for the five-seed headline, a 95\% seed-paired bootstrap interval on $\Delta$ (an uncertainty estimate, not a significance test); we treat a result as reliable only where these agree.

\begin{table*}[t]
\centering
\small
\begin{tabular}{@{}lcccc@{}}
\toprule
\textbf{Variant} & $\alpha$ & $\beta_T$ & $\beta_S$ & Description\\
\midrule
DAPO     & --   & --   & --  & Vanilla DAPO-GRPO; no teacher/student split \\
$\alpha=0$        & 0    & 0  & 0 & Strip-and-retrain (student-only reward maximisation)\\
$\alpha=0.5$      & 0.5  & 0  & 0  & Joint blend of teacher and student reward maximisation \\
$\alpha=1$        & 1    & 0    & 0  & Teacher-only  \\
$\alpha=0.5, \beta$ & 0.5 & $\{0.05,0.1\}$ & $\{0.05,0.1\}$ & Joint blend $+$ symmetric KL (both directions) \\
Filtered SD    & 1    & 0    & $\{0.05,0.1,0.5\}$  & Teacher-only, forward KL toward teacher on filtered samples \\
\quad Unfiltered SD & 1    & 0    & 0.1  & As above, without the correctness filter \\
\bottomrule
\end{tabular}
\caption{Method variants.}
\label{tab:variants}
\end{table*}

\section{Results}
We present results in four parts. \S\ref{sec:results_prelim} sweeps method variants at 1.7B/4K, first with a single seed and then with progressively more seeds for the most promising configurations. \S\ref{sec:results_alpha1_betaS01} reports the headline five-seed comparison establishing ICE with the filtered configuration as the strongest at 1.7B/4K. \S\ref{sec:results_replication} tests whether the gain transfers to longer responses (1.7B/8K) and a larger model (4B/4K). \S\ref{sec:results_passk} separates per-sample accuracy from coverage across sampling budgets for pass@$k$ up to $k=256$.

\subsection{Preliminary Experiments at 1.7B/4K}
\label{sec:results_prelim}
A broad single-seed sweep over $(\alpha, \beta_T, \beta_S)$ (Table~\ref{tab:v3_tier_preliminary}, Appendix~\ref{apdx:prelim-sweep}) identifies $\alpha{=}0.5\,(\beta{=}0.1)$, $\alpha=1$, and the filtered $\beta_S{=}0.1$ configuration as the promising configurations. It also shows that correctness filtering is essential: the unfiltered forward-KL variant falls $12.9\%$ below DAPO. A student reward term ($\alpha<1$) instead makes training unstable, with rising truncation rates (Appendix~\ref{apdx:alpha-dynamics}). These findings motivate the teacher-only, correctness-filtered design. We carry the promising configurations to the multi-seed evaluations below.

\begin{table*}[t]
\centering
\caption{\textbf{Three-seed evaluations:} $\alpha\in\{0,1\}$ and best-performing $\beta$ configurations. $\Delta$ is vs DAPO for method rows and vs Base for the DAPO row.}
\label{tab:v3_tier_baselines}
\small
\begin{tabular}{l c c c c}
\toprule
Method & Held-out & $\Delta$ & $\Delta\%$ & Wins \\
\midrule
Base (pretrain) & $0.201 \pm 0.004$ & --- & --- & --- \\
\midrule
DAPO & $0.245 \pm 0.004$ & $+0.045$ & $+22.4\%$ & --- \\
\midrule
$\alpha{=}0$ & $0.247 \pm 0.005$ & $+0.001$ & $+0.5\%$ & 3/5 \\
$\alpha{=}0.5,\,\beta=0.1$ & $0.249 \pm 0.015$ & $+0.004$ & $+1.5\%$ & 3/5 \\
$\alpha{=}1$ & $0.259 \pm 0.010$ & ${+0.013}$ & ${+5.4\%}$ & 4/5 \\
Filtered SD, $\beta_S{=}0.1$ & $0.261 \pm 0.003$ & ${+0.016}$ & ${+6.5\%}$ & 5/5 \\
\bottomrule
\end{tabular}
\end{table*}
Repeating the most promising configurations across three seeds (Table \ref{tab:v3_tier_baselines}) shrinks several of the single-seed effects. The $\alpha=0.5, \beta_T=\beta_S=0.1$ setting falls from $+0.024$ to $+0.004$ with high variance ($\pm 0.015$). The $\alpha=1$ result decreases from $+0.029$ to $+0.013$ and 4/5 wins. The filtered configuration holds at $+0.016$, 5/5 wins. We carry the two latter configurations forward.

\subsection{The Filtered Configuration Is Strongest at 1.7B/4K}
\label{sec:results_alpha1_betaS01}
\begin{table*}[t]
\centering
\caption{\textbf{Five-seed evaluations:} $\alpha{=}1$ and Filtered SD ($\beta_S{=}0.1$). $\Delta$ is vs DAPO for method rows and vs Base for the DAPO row; CI is a 95\% seed-paired percentile bootstrap on $\Delta$ (10{,}000 resamples). Per-benchmark breakdown in Appendix~\ref{apdx:per-bench}.}
\label{tab:v3_tier_headline}
\small
\begin{tabular}{l c c c c}
\toprule
Method & Held-out & $\Delta$ [95\% CI] & $\Delta\%$ & Wins \\
\midrule
Base (pretrain) & $0.201 \pm 0.004$ & --- & --- & --- \\
\midrule
DAPO & $0.244 \pm 0.004$ & $+0.043$ & $+21.5\%$ & --- \\
\midrule
$\alpha{=}1$ & $0.248 \pm 0.016$ & $+0.005\;[-0.006,\,+0.018]$ & $+1.9\%$ & 4/5 \\
Filtered SD, $\beta_S{=}0.1$ & $0.256 \pm 0.008$ & $+0.012\;[+0.006,\,+0.016]$ & $+5.0\%$ & 5/5 \\
\bottomrule
\end{tabular}
\end{table*}

Extending $\alpha=1$ and the filtered $\beta_S=0.1$ configuration to five seeds (Table \ref{tab:v3_tier_headline}), the $\alpha=1$ improvement reduces further to $+0.005$ ($+1.9\%$) with a CI spanning zero ($[-0.006,+0.018]$), although 4/5 benchmark wins are retained. The filtered configuration is robust at $+0.012$ ($+5.0\%$, $[+0.006,+0.016]$), 5/5 wins.
While inclusion of student policy updates through the forward-KL term improves mean pass@1 by $+0.008$, the 95\% CI marginally spans zero ($[-0.003,+0.016]$; per-benchmark breakdown in Appendix~\ref{apdx:head2head}).

\begin{figure}[t]
\centering
\includegraphics[width=\columnwidth]{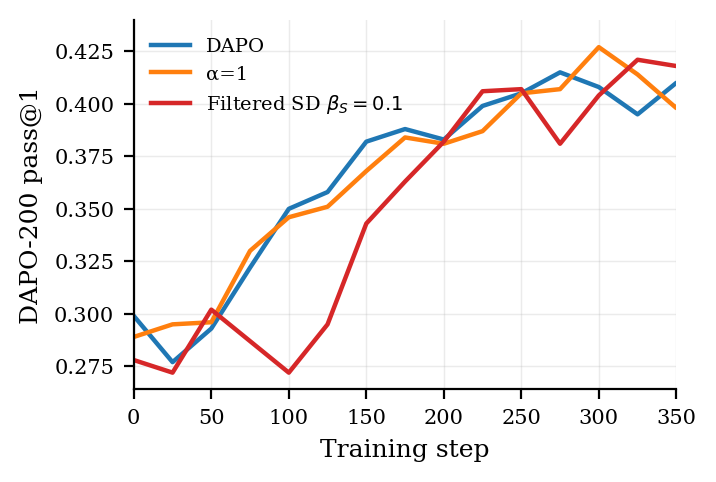}
\caption{DAPO-200 probe pass@1 during training at 1.7B / 4K. Curves are means across $n{=}5$ seeds per method.}
\label{fig:probe_curves}
\end{figure}

Instruction conditioning does not delay final convergence. The filtered configuration lags in the middle of training but catches up by step 200, and the leading 1.7B/4K configurations end level with DAPO (Figure~\ref{fig:probe_curves}; Appendix~\ref{sec:ice_doesnt_slow}).

\subsection{Replication at Longer Response Length and Larger Model Scale}
\label{sec:results_replication}

\begin{table*}[t]
\centering
\caption{\textbf{Replication at 1.7B/8K and 4B/4K.} Held-out is the macro-mean pass@1 over the five benchmarks; $\Delta$ is vs DAPO for method rows and vs Base for the DAPO row. $\alpha{=}1$ was not run at 1.7B/8K. Per-seed values in Appendix~\ref{apdx:per-seed}; per-benchmark breakdown in Appendix~\ref{apdx:per-bench}.}
\label{tab:v3_tier_replication}
\small
\begin{tabular}{l cccc cccc}
\toprule
 & \multicolumn{4}{c}{1.7B / 8K} & \multicolumn{4}{c}{4B / 4K} \\
\cmidrule(lr){2-5}\cmidrule(lr){6-9}
Method & Held-out & $\Delta$ & $\Delta\%$ & Wins & Held-out & $\Delta$ & $\Delta\%$ & Wins \\
\midrule
Base (pretrain) & $0.202 \pm 0.004$ & --- & --- & --- & $0.280 \pm 0.003$ & --- & --- & --- \\
\midrule
DAPO & $0.287 \pm 0.018$ & $+0.084$ & $+41.6\%$ & --- & $0.361 \pm 0.007$ & $+0.081$ & $+28.9\%$ & --- \\
\midrule
$\alpha{=}1$ & --- & --- & --- & --- & $0.358 \pm 0.014$ & $-0.003$ & $-0.9\%$ & 2/5 \\
Filtered SD & $0.297 \pm 0.021$ & $+0.010$ & $+3.5\%$ & 3/5 & $0.345 \pm 0.017$ & $-0.016$ & $-4.5\%$ & 1/5 \\
\bottomrule
\end{tabular}
\end{table*}

We replicate the filtered configuration at 1.7B/8K and 4B/4K, and $\alpha=1$ at 4B/4K (Table \ref{tab:v3_tier_replication}), with 3 seeds per cell. The filtered configuration at 1.7B/8K is the only setting to exceed DAPO ($+0.010$, $+3.5\%$; per-benchmark breakdown in Appendix~\ref{apdx:per-bench}). At 4B/4K neither configuration improves on DAPO: $\alpha=1$ scores $-0.003$ and the filtered configuration $-0.016$ ($-4.5\%$), with two of its three seeds falling below the DAPO mean and one above. The improvement over DAPO observed at 1.7B/4K, and to a lesser extent at 1.7B/8K, does not transfer to the larger model under the 4K-token response limit.

\subsection{Accuracy and Coverage Across Sampling Budgets}
\label{sec:results_passk}
\begin{figure}[t]
\centering
\includegraphics[width=\columnwidth]{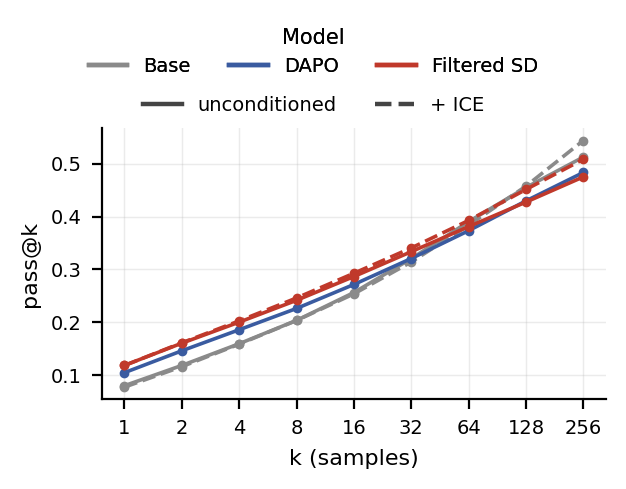}
\caption{Held-out pass@$k$ at 1.7B (macro-mean over AIME-24/25 and HMMT-Feb/Nov) for Base, DAPO, and Filtered SD, sampled unconditioned (solid) and with the behavioural instructions ($+$ICE, dashed). RL post-training leads at low $k$ but trails the base model's coverage at high $k$, which $+$ICE recovers. Per-benchmark breakdown in Appendix~\ref{apdx:passk-breakdown}.}
\label{fig:passk_heldout}
\end{figure}

Sampling at increasing budgets separates per-sample accuracy (pass@1) from coverage (pass@$k$ at large $k$; Figure~\ref{fig:passk_heldout}, Table~\ref{tab:passk}). The filtered configuration leads at pass@1, but at high $k$ the ordering inverts: at 1.7B, RL post-training trades coverage for accuracy, leaving both DAPO and the filtered configuration below the base model at pass@256 ($0.483$ and $0.475$ vs $0.512$). It thus improves pass@1 over DAPO at comparable coverage. Sampling the same checkpoint with the behavioural instructions ($+$ICE) recovers most of this coverage (pass@256 $0.508$). Additionally, the instructions lift even the untrained 1.7B base model's coverage (pass@256 $0.544$ vs $0.512$), indicating that they surface correct solutions independently of post-training.

At 4B the picture reverses. The instructions reduce base-model coverage (pass@256 $0.558$ vs $0.633$), and post-training does not reduce coverage at this response length, with DAPO and the filtered configuration matching or exceeding the base model at pass@256 ($0.642$ vs $0.633$). Truncation rates explain the latter: $21.6\%$ of 4B base-model samples exceed the 4K response limit, against $1.2\%$ after DAPO training (Appendix~\ref{apdx:truncation}). At 4B/4K the binding constraint on coverage is therefore the response budget rather than behavioural support, and the instructions have no coverage headroom to compose, consistent with the absence of a pass@1 gain in \S\ref{sec:results_replication}.

\begin{table}[t]
\centering
\caption{Held-out pass@$k$ at 1.7B and 4B over the four AIME/HMMT benchmarks (MATH-500 excluded), unbiased estimator \citep{chen_evaluating_2021}, $T{=}1.0$, $256$ samples from a single probe-best checkpoint. ``$+$ICE'' samples with the behavioural instructions; this $T{=}1.0$/four-benchmark setup is not directly comparable to Table~\ref{tab:v3_tier_headline}.}
\label{tab:passk}
\small
\begin{tabular}{l c c c}
\toprule
Model & pass@1 & pass@16 & pass@256 \\
\midrule
\multicolumn{4}{l}{\emph{1.7B}} \\
Base         & $0.080$ & $0.256$ & $0.512$ \\
\quad $+$ICE & $0.077$ & $0.254$ & $0.544$ \\
DAPO         & $0.104$ & $0.271$ & $0.483$ \\
Filtered SD  & $0.118$ & $0.286$ & $0.475$ \\
\quad $+$ICE & $0.118$ & $0.293$ & $0.508$ \\
\midrule
\multicolumn{4}{l}{\emph{4B}} \\
Base         & $0.144$ & $0.367$ & $0.633$ \\
\quad $+$ICE & $0.141$ & $0.361$ & $0.558$ \\
DAPO         & $0.229$ & $0.452$ & $0.642$ \\
Filtered SD  & $0.215$ & $0.444$ & $0.642$ \\
\quad $+$ICE & $0.214$ & $0.442$ & $0.617$ \\
\bottomrule
\end{tabular}
\end{table}

\section{Discussion}
We have shown that RL on an instruction-conditioned teacher policy combined with forward-KL distillation to an unconditioned student improves mathematical reasoning over DAPO at 1.7B/4K, with 5/5 benchmark wins and a 95\% CI excluding zero. The improvement over teacher-conditioned RL alone is smaller and less certain. A smaller gain persists at the longer-context 1.7B/8K setting ($+3.5\%$, 3/5 wins), and at 4B/4K it does not improve on DAPO ($-4.5\%$, 1/5 wins). Taken together, these results demonstrate that instruction-conditioned exploration, paired with an explicit self-distillation transfer mechanism, can improve a small model's held-out reasoning over the DAPO baseline. The method changes only training, adding a forward pass per update for the student but no extra rollouts. At test time the student is used exactly as a standard model, so its gains come at no extra inference cost, making it well suited to applications where small models are most useful, such as cost- or latency-constrained deployment.

The improvement from adding the forward-KL term to ICE with RL (filtered $\beta_S{=}0.1$ vs $\alpha=1$) suggests that parameter sharing alone is insufficient to incorporate teacher-sampled behaviour into the student policy; the explicit distillation signal contributes additional transfer. Another possible effect is that because student and teacher share parameters, the KL-driven updates also affect the teacher's subsequent rollouts, producing a compounding effect beyond what the RL objective alone can produce.

Beyond pass@1, the budgeted sampling in \S\ref{sec:results_passk} ties these gains to exploration: at 1.7B, RL post-training raises pass@1 but reduces high-$k$ coverage, the \textit{support shrinkage} of \citet{wu_invisible_2026}, while sampling with the behavioural instructions restores coverage to near the base model, realising the \textit{support composition} they induce and offsetting, on this proxy, the exploration that post-training removes.

At 4B/4K the gain seen at 1.7B does not re-emerge: ICE with the filtered configuration does not improve over DAPO ($-4.5\%$, 1/5 wins), even though the same mechanism delivers a clear gain at 1.7B. The point estimate is negative, but high seed variance ($\sigma{=}0.017$) leaves the 95\% CI $[-0.036, +0.008]$ spanning zero. The coverage analysis of \S\ref{sec:results_passk} explains the difference. At 1.7B the instructions expand the support of the base policy (pass@256 $0.544$ vs $0.512$), providing behaviours for the self-distillation to transfer, whereas at 4B they reduce it ($0.558$ vs $0.633$), leaving no coverage headroom to compose. Comparing base-model pass@$k$ with and without instructions therefore provides a cheap, training-free test of whether ICE can help in a given setting.

Although the instructions were selected using the 4B model, the selection procedure measured coverage at small sampling budgets on the easier training-distribution problems; this does not imply an expanded reachable set on the harder held-out problems. Additionally, we test 4B only at the 4K response length, where truncation is frequent ($21.6\%$ of base-model samples, Appendix~\ref{apdx:truncation}), so its behaviour at longer contexts is unknown.

Because ICE's instructions are fixed and untargeted, a more efficient approach to generating instructions could improve sample efficiency (although Appendix~\ref{sec:ice_doesnt_slow} does show convergence is not slowed by including broad, potentially unhelpful instructions). This motivates generating bespoke, per-task strategies dynamically, as in the self-feedback methods of \S\ref{sec:related_context_explore} such as Strategy-Guided Exploration \citep{szot_expanding_2026}, optionally drawing on privileged information such as the correct answer.

\section{Conclusions}
We have introduced instruction-conditioned exploration (ICE), a template-based context-mediated exploration method that conditions RL post-training on a fixed set of diverse behavioural instructions, and combined it with filtered off-policy self-distillation to transfer instruction-conditioned behaviour into an unconditioned test-time policy.

At 1.7B/4K, RL on the instruction-conditioned teacher combined with forward-KL distillation to the unconditioned student improves held-out pass@1 on mathematical reasoning tasks over DAPO with 5/5 benchmark wins and a 95\% CI excluding zero. The gain holds across both the 4K and 8K context lengths at 1.7B (more weakly at 8K), suggesting it is not specific to a single setting, though it does not appear at 4B/4K, where the instructions do not expand base-model coverage (\S\ref{sec:results_passk}). These results demonstrate that instruction-conditioning during training, combined with self-distillation, can improve the reasoning of small language models.

We intend the 1.7B/8K and 4B/4K results as first steps toward characterising the approach across a range of small models; future work should build on them with additional seeds, model sizes, context lengths and task domains to map where the gain holds, targeting settings where the base-model coverage test indicates that the instructions add headroom.

\section{Limitations}

\paragraph{Statistics and evaluation.} Seed counts are low: headline configurations use $n{=}3$--$5$ seeds and several preliminary variants are single-seed, so our strongest result ($\alpha{=}1,\beta_S{=}0.1$, filtered) should be replicated at higher $n$ before strong conclusions are drawn. These configurations were also chosen from a single-seed sweep, so the selection seed inflates their headline estimates, which are therefore mildly optimistic. The held-out benchmarks compound this: AIME-24/25 and HMMT-Feb/Nov-25 contain only 30 questions each and carry substantial pass@1 noise, and because these pairs measure overlapping skills the effective number of independent benchmarks is below five, so benchmark-win counts overstate significance when read alone. We therefore treat wins as one signal among several, alongside the mean improvement and its bootstrap interval, rather than as a standalone significance test. Held-out scores are reported at the step maximising a 200-question training-set probe; this is identical across methods, keeping comparisons fair, though absolute numbers may be optimistic. Finally, MATH-500 predates Qwen3's pretraining cut-off and was likely seen during pretraining; we retain it for comparability with prior work but it should be read with this in mind.

\paragraph{Baselines and scope.} The only baseline is vanilla DAPO; we do not compare directly against other context-mediated exploration methods, in particular Prompt Augmentation \citep{lu_prompt_2026}, the closest in spirit to ICE, so gains are measured over an unconditioned RL baseline rather than alternative exploration approaches. The study is also narrow: a single domain (mathematical reasoning with binary verifiable rewards), one base-model family (Qwen3 with thinking disabled), two model scales and two context lengths, a sparse sweep of $(\alpha,\beta_T,\beta_S)$, and a fixed training budget. Generalisation to other domains, algorithms, and larger models therefore remains open.

\paragraph{Method.} ICE requires a small set of describable behavioural instructions, supplied by an expert or a capable model, and samples them without per-problem targeting; adaptive, per-task alternatives are left to future work. Potential risks of the work are discussed in Appendix~\ref{apdx:risks}.

\section{Acknowledgements}

The authors acknowledge the financial support from the Engineering and Physical Sciences Research Council (EPSRC) through a Doctoral Training Partnership (DTP) grant (EP/W524621/1) and a Turing AI Fellowship on `Citizen-Centric AI Systems' (EP/V022067/1).

The authors acknowledge the use of resources provided by the Isambard-AI National AI Research Resource (AIRR). Isambard-AI is operated by the University of Bristol and is funded by the UK Government's Department for Science, Innovation and Technology (DSIT) via UK Research and Innovation; and the Science and Technology Facilities Council [ST/AIRR/I-A-I/1023].

The authors acknowledge the use of the IRIDIS High Performance Computing Facility, and associated support services at the University of Southampton, in the completion of this work.

\bibliography{references}

\appendix

\section{Supplementary Instructions}
\label{sec:appendix-instructions}

We design an automated and replicable process for selecting behavioural instructions. Claude Sonnet 4.6 solves a 250-problem subset of the training set, with Claude Opus 4.6 retrying any problems it answers incorrectly. The correct solutions are categorised and synthesised into 10 candidate strategies, each with 3 instruction-text variants. The model to be post-trained (Qwen3 4B) then attempts a separate held-out 250-problem subset under every strategy and variant, scored by pass@16. We select the 5-strategy set by greedy set cover on pass@16 coverage: at each step we add the strategy that solves the most as-yet-unsolved problems, breaking ties by per-sample success rate. 
One variant per selected strategy is used as its instruction, yielding the $N=5$ instructions in Table \ref{tab:instructions}. The coverage provided by the instruction set is insensitive to this selection procedure (Appendix~\ref{apdx:subset-sensitivity}).

\begin{table*}[t]
\centering
\small
\begin{tabular}{@{}lp{0.78\linewidth}@{}}
\toprule
\textbf{ID} & \textbf{Instruction text} \\
\midrule
\texttt{work\_backwards} & ``Invert the process: ask what input would produce this output, rather than computing forward.'' \\
\addlinespace[2pt]
\texttt{constructive\_building} & ``After proving a bound, construct a concrete configuration that achieves it to confirm tightness.'' \\
\addlinespace[2pt]
\texttt{symmetry\_and\_invariance} & ``Ask: what symmetry does this problem have, and how can it reduce the number of cases or simplify the expression?'' \\
\addlinespace[2pt]
\texttt{decomposition} & ``Identify which components of the problem can be handled independently and tackle them one at a time.'' \\
\addlinespace[2pt]
\texttt{reduction} & ``Reformulate the problem in a different domain where the answer becomes obvious or follows from a standard result.'' \\
\bottomrule
\end{tabular}
\caption{The set of behavioural instructions $\mathcal{I}$ used during training.}
\label{tab:instructions}
\end{table*}

\section{Instruction-Set Sensitivity}
\label{apdx:subset-sensitivity}
The greedy set-cover selection of Appendix~\ref{sec:appendix-instructions} is not critical to the instruction set used. Table~\ref{tab:subset_sensitivity} reports the combined coverage of every possible five-strategy subset of the ten-strategy candidate pool, where a problem is covered if at least one of the $16$ samples from any of the subset's instruction variants solves it (three variants per strategy, as in the selection procedure). All $\binom{10}{5}=252$ subsets fall within $8$ problems ($3.2\%$ of the subset) of the optimum, and $81\%$ fall within $5$ problems. Within this candidate pool, coverage is insensitive to which five strategies are selected.

\begin{table}[t]
\centering
\small
\begin{tabular}{@{}lcc@{}}
\toprule
\textbf{Instruction subset} & \textbf{Coverage} & \textbf{\% of subset} \\
\midrule
Greedy-selected set (selected) & $235$ & $94.0\%$ \\
Best of all five-strategy subsets & $235$ & $94.0\%$ \\
Median subset & $231$ & $92.4\%$ \\
Worst subset & $227$ & $90.8\%$ \\
\bottomrule
\end{tabular}
\caption{Sensitivity of coverage to instruction-set selection. Combined coverage (problems solved by at least one instruction variant of a strategy in the subset, $16$ samples per variant, Qwen3-4B) over the 250-problem selection subset, for all $\binom{10}{5}=252$ five-strategy subsets of the candidate pool.}
\label{tab:subset_sensitivity}
\end{table}

\section{Compute}
\label{apdx:compute}

Training was conducted with NVIDIA GH200 superchips on the Isambard-AI cluster \citep{mcintosh-smith_isambard-ai_2024}. Each 1.7B configuration used 2 GPUs requiring approximately 100--250 GPU-hours per seed. Each 4B configuration used 4 GPUs requiring approximately 200--250 GPU-hours per seed. Training with the filtered configuration requires approximately 1.35 times the GPU-time per training step of DAPO on identical hardware, with rollout time essentially unchanged; the additional cost is the student forward pass in the update. The experiments reported in this work total approximately $6{,}500$ GPU-hours. Including all preliminary and exploratory work (pilots, infrastructure development, and runs not included in the final results), total project compute amounted to roughly $10{,}000$ GPU-hours.

\section{Artifact Licenses}
\label{apdx:licenses}
All artifacts used in this work are openly released and were used for research purposes consistent with their licenses. The Qwen3 models \citep{yang_qwen3_2025} and the DAPO-Math-17K dataset \citep{yu_dapo_2025} are released under the Apache-2.0 license, as is the OpenRLHF framework used for training. For evaluation, MATH-500 (a subset of the MATH dataset \citep{hendrycks_measuring_2021} distributed through OpenAI's PRM800K release) is under the MIT license, the AIME-24/25 sets \citep{zhang_american_2024, zhang_american_2025} are under Apache-2.0, and the HMMT-Feb/Nov-25 sets obtained via MathArena \citep{balunovic_matharena_2026} are under CC BY-NC-SA 4.0. Our use is non-commercial academic research, consistent with these terms. The behavioural instructions and source code produced are released for research use only.

\section{Potential Risks}
\label{apdx:risks}
Our method improves the mathematical reasoning of small language models through a training-time-only intervention, and we foresee limited direct risk. As with any technique that raises the capability of compact, cheaply deployable models, the gains are broadly dual-use: more capable small models lower the barrier to both beneficial and potentially harmful downstream applications. Our experiments are confined to mathematical reasoning with binary verifiable rewards, a domain with little scope for harmful content, and we release only behavioural instructions and source code intended for research use. We introduce no new data collection involving people, and the datasets used contain no personal or sensitive information.

\section{Preliminary Single-Seed Sweep}
\label{apdx:prelim-sweep}
We first run a broad preliminary sweep with a single seed per configuration, then repeat the most promising with additional seeds.

\begin{table*}[t]
\centering
\caption{\textbf{Preliminary results, single-seed.} $\Delta$ is vs DAPO for method rows and vs Base for the DAPO row.}
\label{tab:v3_tier_preliminary}
\small
\begin{tabular}{l c c c c}
\toprule
Method & Held-out & $\Delta$ & $\Delta\%$ & Wins \\
\midrule
Base (pretrain) & $0.201 \pm 0.004$ & --- & --- & --- \\
\midrule
DAPO & $0.241$ & $+0.041$ & $+20.4\%$ & --- \\
\midrule
$\alpha{=}0$ & $0.244$ & $+0.003$ & $+1.1\%$ & 3/5 \\
$\alpha{=}0.5$ & $0.238$ & $-0.004$ & $-1.5\%$ & 2/5 \\
$\alpha{=}0.5,\,\beta{=}0.05$ & $0.242$ & $+0.001$ & $+0.2\%$ & 2/5 \\
$\alpha{=}0.5,\,\beta{=}0.1$ & $0.266$ & $+0.024$ & $+10.1\%$ & 5/5 \\
$\alpha{=}1$ & $0.270$ & $+0.029$ & $+12.0\%$ & 5/5 \\
Unfiltered SD, $\beta_S{=}0.1$ & $0.210$ & $-0.031$ & $-12.9\%$ & 0/5 \\
Filtered SD, $\beta_S{=}0.05$ & $0.243$ & $+0.001$ & $+0.5\%$ & 4/5 \\
Filtered SD, $\beta_S{=}0.1$ & $0.258$ & $+0.017$ & $+7.0\%$ & 3/5 \\
Filtered SD, $\beta_S{=}0.5$ & $0.208$ & $-0.034$ & $-14.0\%$ & 0/5 \\
\bottomrule
\end{tabular}
\end{table*}
Three configurations show promise in the single-seed sweep at 1.7B/4K (Table \ref{tab:v3_tier_preliminary}): $\alpha=0.5, \beta=0.1$ ($+0.024$, 5/5 wins), $\alpha=1$ ($+0.029$, 5/5), and the filtered $\beta_S=0.1$ configuration ($+0.017$, 3/5). The $\alpha\in\{0,0.5\}$ configurations without KL regularisation give no improvement over DAPO; these configurations, which include a student reward term, additionally exhibit increased truncation rates during training (Appendix \ref{apdx:alpha-dynamics}).
Setting $\beta_S$ to $0.5$ in the filtered configuration degrades performance considerably ($-0.034$, 0/5) relative to $\beta_S=0.1$, indicating that the forward-KL weight requires careful tuning: $\beta_S=0.1$ is the best option tested.

Filtering on correctness is critical for the $\alpha=1, \beta_S>0$ setting: the unfiltered variant decreases mean pass@1 by $-0.031$ ($-12.9\%$) relative to the baseline, with worse performance on all 5 benchmarks. Including the forward-KL objective trains the student to imitate the teacher; incorrect rollouts therefore yield harmful student updates.

\section{Head-to-Head: Filtered $\beta_S{=}0.1$ vs.\ $\alpha{=}1$}
\label{apdx:head2head}
Table~\ref{tab:v3_owf_vs_alpha1_probe} reports the per-benchmark paired comparison between the filtered $\beta_S{=}0.1$ configuration and $\alpha{=}1$ at 1.7B/4K referenced in \S\ref{sec:results_alpha1_betaS01}.

\begin{table}[t]
\centering
\caption{Head-to-head paired bootstrap (as in Table~\ref{tab:v3_tier_headline}) $\Delta = \big(\text{Filtered SD, $\beta_S=0.1$}\big) - \big(\alpha{=}1\big)$ at 1.7B / 4K. $n{=}5$ paired training seeds.}
\label{tab:v3_owf_vs_alpha1_probe}
\small
\begin{tabular}{l c c}
\toprule
Benchmark & $\Delta$ [95\% CI] & $\Delta\%$ \\
\midrule
MATH-500 & $+0.008\;[-0.016,\,+0.025]$ & $+1.0\%$ \\
AIME 24 & $+0.016\;[-0.011,\,+0.043]$ & $+9.2\%$ \\
AIME 25 & $+0.019\;[+0.003,\,+0.035]$ & $+11.9\%$ \\
HMMT Feb & $-0.013\;[-0.032,\,+0.002]$ & $-17.2\%$ \\
HMMT Nov & $+0.009\;[+0.000,\,+0.017]$ & $+19.6\%$ \\
\midrule
Held-out avg & $+0.008\;[-0.003,\,+0.016]$ & $+3.1\%$ \\
\midrule
Wins & 4/5 & --- \\
\bottomrule
\end{tabular}
\end{table}

\section{Per-Benchmark Held-out Results}
\label{apdx:per-bench}
Table~\ref{tab:v3_per_bench} breaks the held-out macro-means of Tables~\ref{tab:v3_tier_headline} and~\ref{tab:v3_tier_replication} down by benchmark. At 1.7B/8K the two benchmarks where the filtered configuration trails DAPO are the HMMT pair, with margins ($-0.005$ and $-0.003$) well within the per-benchmark noise of these 30-question sets.

\begin{table*}[t]
\centering
\caption{Per-benchmark held-out pass@1 (probe-best selection; mean$\pm$std across training seeds). $\Delta$ is Filtered SD $-$ DAPO; \textbf{bold} marks benchmarks where Filtered SD trails DAPO. Macro-means (Held-out avg) reproduce Tables~\ref{tab:v3_tier_headline} and~\ref{tab:v3_tier_replication}. Filtered SD wins 5/5 at 1.7B/4K, 3/5 at 1.7B/8K (trailing on the two HMMT sets), and 1/5 at 4B/4K.}
\label{tab:v3_per_bench}
\small
\setlength{\tabcolsep}{4pt}
\begin{tabular}{l c c c c c c c}
\toprule
Method & n & MATH-500 & AIME 24 & AIME 25 & HMMT Feb & HMMT Nov & Held-out avg \\
\midrule
\multicolumn{8}{l}{\textbf{1.7B / 4K}} \\
DAPO & 5 & $0.776 \pm 0.009$ & $0.164 \pm 0.015$ & $0.170 \pm 0.015$ & $0.062 \pm 0.015$ & $0.046 \pm 0.016$ & $0.244 \pm 0.004$ \\
Filtered SD & 5 & $0.793 \pm 0.014$ & $0.187 \pm 0.028$ & $0.180 \pm 0.020$ & $0.064 \pm 0.018$ & $0.056 \pm 0.006$ & $0.256 \pm 0.008$ \\
$\Delta$ (Filtered SD $-$ DAPO) & & $+0.017$ & $+0.023$ & $+0.009$ & $+0.003$ & $+0.010$ & $+0.012$ \\
\addlinespace
\multicolumn{8}{l}{\textbf{1.7B / 8K}} \\
DAPO & 3 & $0.813 \pm 0.023$ & $0.220 \pm 0.040$ & $0.208 \pm 0.020$ & $0.111 \pm 0.017$ & $0.081 \pm 0.003$ & $0.287 \pm 0.018$ \\
Filtered SD & 3 & $0.831 \pm 0.023$ & $0.241 \pm 0.027$ & $0.226 \pm 0.028$ & $0.106 \pm 0.020$ & $0.078 \pm 0.019$ & $0.297 \pm 0.021$ \\
$\Delta$ (Filtered SD $-$ DAPO) & & $+0.019$ & $+0.021$ & $+0.018$ & $\mathbf{-0.005}$ & $\mathbf{-0.003}$ & $+0.010$ \\
\addlinespace
\multicolumn{8}{l}{\textbf{4B / 4K}} \\
DAPO & 3 & $0.883 \pm 0.006$ & $0.324 \pm 0.007$ & $0.272 \pm 0.021$ & $0.145 \pm 0.010$ & $0.183 \pm 0.016$ & $0.361 \pm 0.007$ \\
Filtered SD & 3 & $0.858 \pm 0.012$ & $0.306 \pm 0.024$ & $0.272 \pm 0.026$ & $0.122 \pm 0.007$ & $0.166 \pm 0.027$ & $0.345 \pm 0.017$ \\
$\Delta$ (Filtered SD $-$ DAPO) & & $\mathbf{-0.025}$ & $\mathbf{-0.017}$ & $+0.001$ & $\mathbf{-0.023}$ & $\mathbf{-0.017}$ & $-0.016$ \\
\addlinespace
\bottomrule
\end{tabular}
\end{table*}

\section{Per-Seed Held-out Results}
\label{apdx:per-seed}
Table~\ref{tab:v3_per_seed_replication} reports the per-seed held-out averages underlying the replication table (\S\ref{sec:results_replication}, Table~\ref{tab:v3_tier_replication}), under probe-best selection. The three 4B/4K filtered-configuration seeds are $0.333$, $0.365$, and $0.337$ (mean $0.345 \pm 0.017$), two below DAPO and one above.

\begin{table}[t]
\centering
\caption{Per-seed held-out pass@1 (macro-mean over MATH-500, AIME 24/25, HMMT Feb/Nov), probe-best selection. ``DAPO'' is the RL-only baseline.}
\label{tab:v3_per_seed_replication}
\small
\setlength{\tabcolsep}{4pt}
\begin{tabular}{l c c c c}
\toprule
Method & Seed 1 & Seed 2 & Seed 3 & Mean$\pm$std \\
\midrule
\multicolumn{5}{l}{\emph{1.7B / 8K}} \\
DAPO            & $0.308$ & $0.276$ & $0.277$ & $0.287 \pm 0.018$ \\
Filtered SD     & $0.304$ & $0.312$ & $0.273$ & $0.297 \pm 0.021$ \\
\midrule
\multicolumn{5}{l}{\emph{4B / 4K}} \\
DAPO            & $0.370$ & $0.357$ & $0.357$ & $0.361 \pm 0.007$ \\
$\alpha{=}1$    & $0.373$ & $0.344$ & $0.357$ & $0.358 \pm 0.014$ \\
Filtered SD     & $0.333$ & $0.365$ & $0.337$ & $0.345 \pm 0.017$ \\
\bottomrule
\end{tabular}
\end{table}

\section{Per-Benchmark pass@$k$}
\label{apdx:passk-breakdown}
\begin{figure}[t]
\centering
\includegraphics[width=\columnwidth]{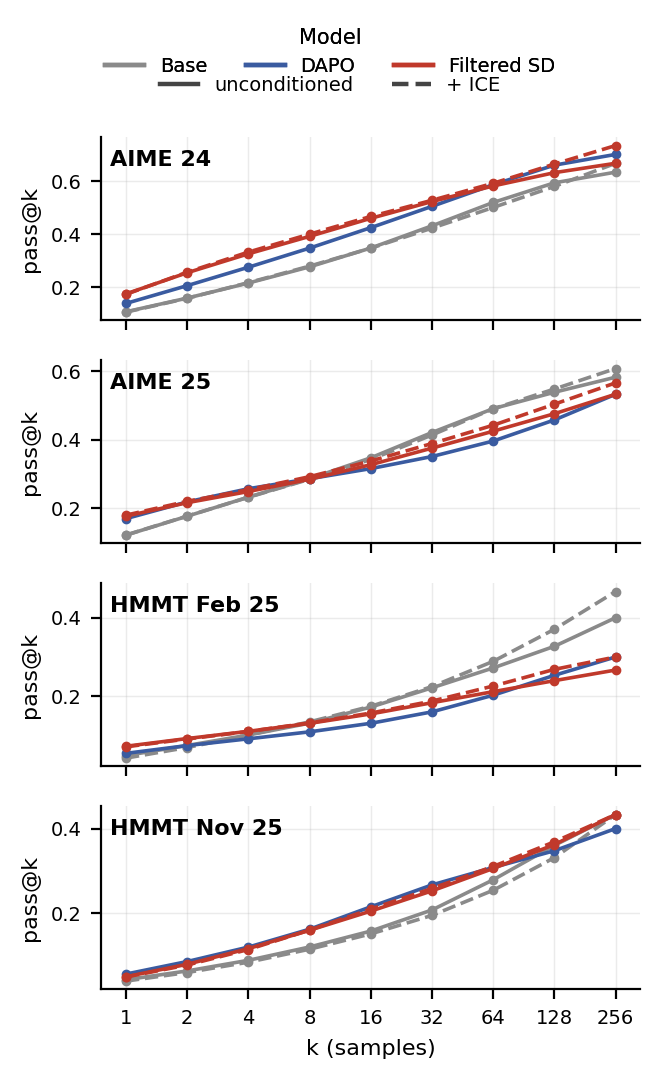}
\caption{Per-benchmark pass@$k$ at 1.7B for the held-out benchmarks (AIME-24/25, HMMT-Feb/Nov), for Base, DAPO, and Filtered SD, sampled unconditioned (solid) and with the behavioural instructions ($+$ICE, dashed).}
\label{fig:passk_stacked}
\end{figure}

Figure~\ref{fig:passk_stacked} breaks the held-out pass@$k$ curves of Figure~\ref{fig:passk_heldout} down by benchmark.

\section{Truncation Rates in Budgeted Sampling}
\label{apdx:truncation}
Table~\ref{tab:truncation} reports the fraction of samples exceeding the 4K response limit in the pass@$k$ evaluations of Table~\ref{tab:passk}.

\begin{table}[t]
\centering
\small
\begin{tabular}{l c c}
\toprule
\textbf{Model} & \textbf{1.7B} & \textbf{4B} \\
\midrule
Base         & $8.5\%$ & $21.6\%$ \\
\quad $+$ICE & $6.6\%$ & $19.6\%$ \\
DAPO         & $0.6\%$ & $1.2\%$ \\
Filtered SD  & $6.0\%$ & $9.7\%$ \\
\quad $+$ICE & $6.8\%$ & $10.2\%$ \\
\bottomrule
\end{tabular}
\caption{Truncation rates (fraction of samples exceeding the 4K response limit) in the pass@$k$ evaluations of Table~\ref{tab:passk}.}
\label{tab:truncation}
\end{table}

\section{Instruction Conditioning Does Not Slow Learning}
\label{sec:ice_doesnt_slow}
Despite replacing some neutral prompts with potentially unhelpful instruction-conditioned ones, instruction conditioning does not slow convergence: the $\alpha=1$ probe pass@1 tracks DAPO throughout training, pulling clearly ahead only late in training (around 250--300 steps; Figure \ref{fig:probe_curves}). The filtered configuration ($\beta_S{=}0.1$) lags at early steps but catches up to $\alpha=1$ by step 200.

\section{Training Dynamics of the $\alpha$-Sweep}
\label{apdx:alpha-dynamics}

\begin{figure}[t]
\centering
\includegraphics[width=\columnwidth]{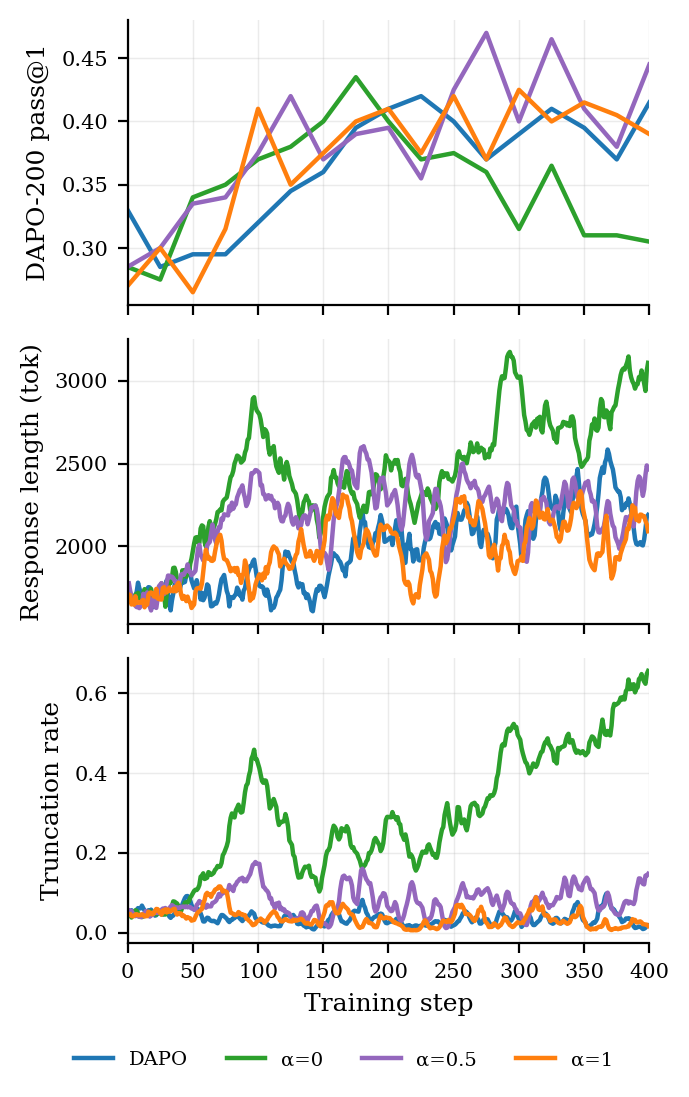}
\caption{$\alpha$-sweep training dynamics at 1.7B / 4K from a single training seed per method.}
\label{fig:alpha_dynamics}
\end{figure}

Configurations with $\alpha\in\{0,0.5\}$, which include a student reward term with importance-sampling correction, exhibit increased truncation rates during training, with $\alpha=0$ also showing response-length growth. The truncation rate for both DAPO and $\alpha=1$ remains low throughout training (Figure \ref{fig:alpha_dynamics}), ending at $0.023$ and $0.020$ respectively (Table \ref{tab:alpha_dynamics_supplement}). In contrast $\alpha=0$ and $\alpha=0.5$ increase to $0.616$ and $0.093$ respectively at step 400. In particular, the response length of $\alpha=0$ reaches $3033$ tokens, $1.40$ times that of DAPO ($2160$ tokens).

\begin{table}[t]
\centering
\caption{Endpoint summary for the $\alpha$-sweep at 1.7B / 4K (DAPO-200 at step 400, response length (number of tokens) and truncation rate are the mean of steps $376$--$400$). Each row is from a single training seed.}
\label{tab:alpha_dynamics_supplement}
\small
\begin{tabular}{l c c c}
\toprule
Method & DAPO-200 & Resp.\ length & Truncation rate \\
\midrule
DAPO & $0.415$ & $2160$ & $0.023$ \\
$\alpha{=}0$ & $0.305$ & $3033$ & $0.616$ \\
$\alpha{=}0.5$ & $0.445$ & $2231$ & $0.093$ \\
$\alpha{=}1$ & $0.390$ & $2114$ & $0.020$ \\
\bottomrule
\end{tabular}
\end{table}

\end{document}